\pdfoutput=1

\documentclass[11pt]{article}
\usepackage[pdfencoding=auto,psdextra]{hyperref}
\usepackage{EMNLP2022}

\usepackage{times}
\usepackage{latexsym}
\usepackage[T1]{fontenc}
\usepackage[utf8]{inputenc}
\usepackage{microtype}

\usepackage{graphicx}
\usepackage{amsmath}
\usepackage{amsthm}
\usepackage{amsfonts}
\usepackage{amssymb}
\usepackage{bm}
\usepackage{booktabs}
\usepackage{multirow}
\usepackage{standalone}
\usepackage{algorithm}
\usepackage{algpseudocode}
\usepackage{makecell}
\usepackage{soul}
\usepackage{xspace}

\usepackage{verbatim}
\usepackage{color}

\newcommand{\eg}{\emph{e.g.,\xspace}}
\newcommand{\ie}{\emph{i.e.,\xspace}}

\DeclareMathOperator*{\argmax}{arg\,max}

\title{Unifying the Convergences in Multilingual Neural Machine Translation}

\author{Yichong Huang \hspace{1em} Xiaocheng Feng \hspace{1em} Xinwei Geng \hspace{1em} Bing Qin\\
   Harbin Institute of Technology, China \\
  \texttt{\{ychuang, xcfeng, xwgeng, qinb\}@ir.hit.edu.cn} \\}

\begin{document}
\maketitle

\begin{abstract}
Although all-in-one-model multilingual neural machine translation (multilingual NMT) has achieved remarkable progress, the \textit{convergence inconsistency} in the joint training is ignored, \ie{different language pairs reaching convergence in different epochs}.
This leads to the trained MNMT model over-fitting low-resource language translations while under-fitting high-resource ones.
In this paper, we propose a novel training strategy named LSSD (\textbf{L}anguage-\textbf{S}pecific \textbf{S}elf-\textbf{D}istillation), which can alleviate the convergence inconsistency and help MNMT models achieve the best performance on each language pair simultaneously.
Specifically, LSSD picks up \textit{language-specific best checkpoints} for each language pair to teach the current model on the fly.
Furthermore, we systematically explore three sample-level manipulations of knowledge transferring. 
Experimental results on three datasets show that LSSD obtains consistent improvements towards all language pairs and achieves the state-of-the-art
\footnote{Our code is publicly available at \url{https://github.com/OrangeInSouth/LSSD}.}
. 

\end{abstract}
\section{Introduction}
\begin{figure}[ht]
  \centering
    \includegraphics[clip,width=1.0\columnwidth,]{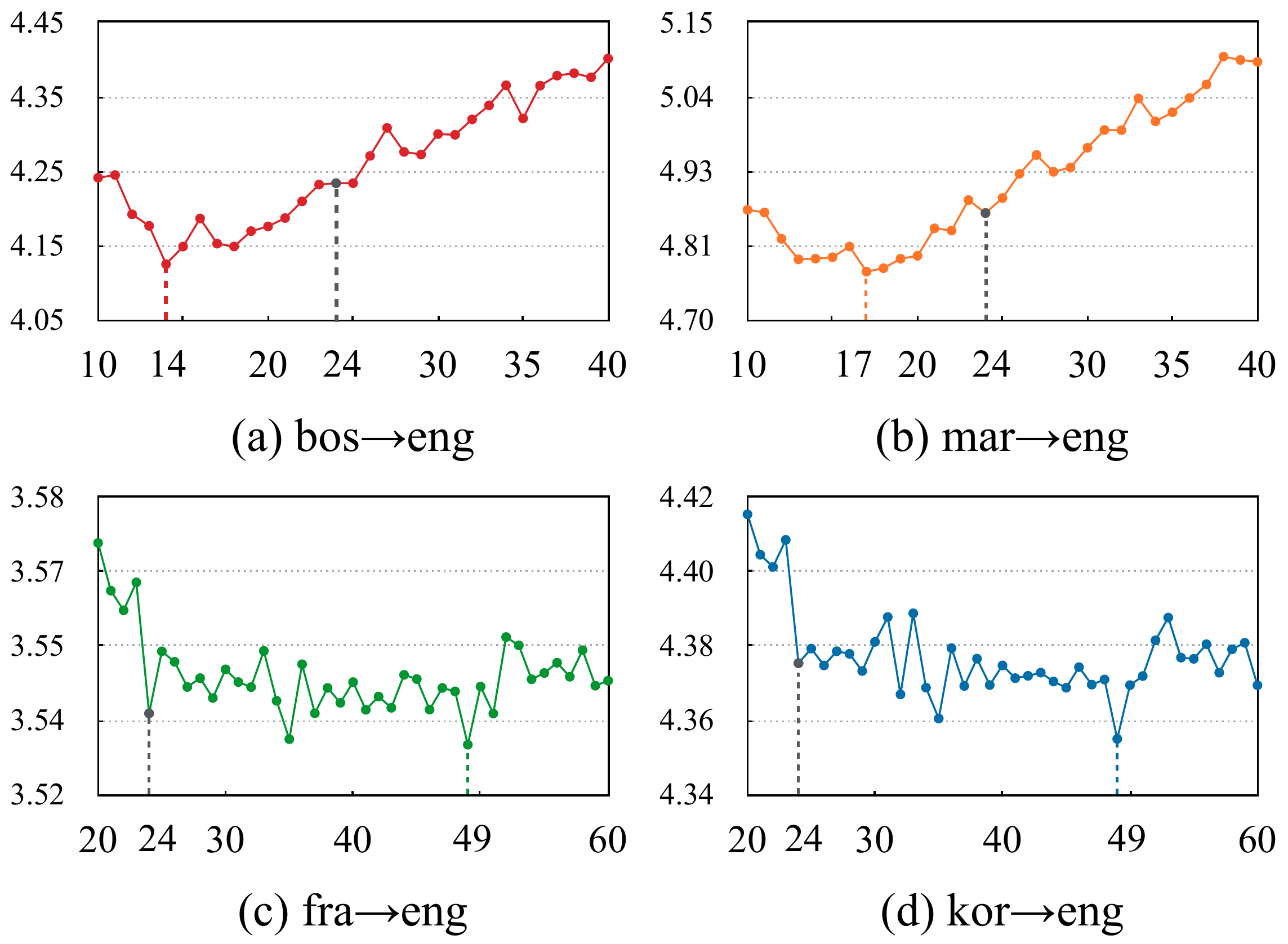}
    \caption{Loss curves of a multilingual NMT model trained to translate 4 languages into English. The upper two boxes illustrate low-resource directions. And the high-resource ones are placed in the bottom part. The x-axis and y-axis indicate training epochs and dev losses respectively. Dotted lines mark language-specific best checkpoints (colorful ones) as well as the overall best checkpoint (the black one).}
  \label{fig:case}
\end{figure}
Neural machine translation (NMT)~\cite{kalchbrenner-blunsom-2013-recurrent,sutskever-etal-2014-sequence} has witnessed enormous and significant progress, including network structures~\cite{cho-etal-2014-learning,bahdanau-etal-2015-neural,gehring-etal-2017-convolutional,vaswani-etal-2017-attention}, attention mechanism~\cite{luong-etal-2015-effective,liu-etal-2016-neural,shen-etal-2018-disan} and decoding strategies~\cite{xia-etal-2017-deliberation,geng-etal-2018-adaptive,zhou-etal-2019-synchronous}.
While achieving promising performance~\cite{wu-etal-2016-googles,hassan-etal-2018-achieving}, widely-used {\em bilingual} NMT actually causes huge computational cost especially when tackling numerous language pairs, thereby facilitating the recent emergence of {\em multilingual} NMT~\cite{ha-etal-2016-toward,firat-etal-2016-multi,johnson-etal-2017-googles,lu-etal-2018-neural,aharoni-etal-2019-massively}. 
By directly translating multiple language pairs with one model, multilingual NMT~\cite{tan-etal-2018-multilingual,zhang-etal-2020-improving,fan-etal-2021-beyond,zhang-etal-2021-share} quadratically accelerates deployment and effectively encourages transfer learning between similar languages, which greatly benefits low-resource directions~\cite{arivazhagan-etal-2019-massively} and successfully enables zero-shot translation~\cite{gu-etal-2019-improved}.

Despite the remarkable success, multilingual NMT~\cite{liu-etal-2020-multilingual-denoising,lin-etal-2021-learning} clearly expresses a strong disagreement on the uniform convergence point across various translation corpora.
The underlying reason is the imbalance and heterogeneity of available data in multilingual training~\cite{wang-etal-2020-balancing,wu-etal-2021-uncertainty}, which also explains why the disagreement between high-resource languages (HRLs) and low-resource languages (LRLs) is more pronounced, as illustrated in Figure~\ref{fig:case}. 
Particularly, a fairly common and important observation about the learning of multilingual NMT is that
HRLs typically encounter underfitting, while extremely severe overfitting generally arises in LRLs. 

In this paper, we cast this convergence inconsistency as the performance deficit between the multilingual NMT model and its own language-specific best checkpoints, and aim to reduce this deficit.
Towards tacking this problem, we propose a novel training strategy dubbed \textbf{L}anguage-\textbf{S}pecific \textbf{S}elf-\textbf{D}istillation (LSSD).
At each training step, LSSD appoints recent best checkpoints oriented to each language pair as teacher models and treats the current training model as the student learning from each teacher model in a knowledge distillation manner.
Differently depending on data selection and strength of performing knowledge distillation indeed, several practical strategies are employed to potentially provide a relatively fine-grained manipulation of knowledge transferring:
1) \textsc{LSSD-Whole}, which performs distillation on all data samples equally; 
2) \textsc{LSSD-Selective}, which selects samples for distillation conditioned on whether the teacher performs better than the student;
3) \textsc{LSSD-Adaptive}, which varies the distillation strength according to the sample-level performance ratio between the teacher and student.

Experimental results on TED talks\footnote{\url{https://www.ted.com/participate/translate}} and WMT under {\em many-to-one} and {\em one-to-many} settings demonstrate that our method can obtain consistent and significant improvement through remedying the convergence inconsistency, ultimately achieving state-of-the-art translation performance.

\section{Preliminaries}
\subsection{Neural Machine Translation}
Bilingual neural machine translation (Bilingual NMT) translates a sentence $x$ in source language into a sentence $y$ in target language. Given a parallel corpus ${D} = \{(x,y) \in \mathcal{X} \times \mathcal{Y}\}$, the neural machine translation model is commonly trained with the Maximum Likelihood Estimation (MLE):
\begin{align}
\theta^* &= \argmax\limits_{\theta} \mathbb{E}_{(x,y) \sim {D}} \sum\limits_{i \le |y|} \log P(y_i | x,y_{<i}; \theta),
\label{eq:MLE_loss}
\end{align}
 where $P(\cdot|\cdot; \theta)$ is the conditional probability with model $\theta$, which is usually implemented in an encoder-decoder architecture~\citep{bahdanau-etal-2015-neural,vaswani-etal-2017-attention}.
 
\subsection{Multilingual Neural Machine Translation}
Multilingual neural machine translation (multilingual NMT) translates multiple language pairs with one unified model.
In this work, we follow \citet{johnson-etal-2017-googles} to train a multilingual NMT model jointly using training datasets of $L$ language pairs $D^{train}=\{D^{train}_1,...,D^{train}_l...,D^{train}_L\}$, where $D^{train}_l$ is the dataset of language pair $(S_l,T_l)$. 
To encode and decode diverse languages to/from a shared semantic space, a large multilingual vocabulary $\mathcal{V}$ is constructed.
And a language tag is appended to the beginning of source sentences to specify the target language.
Similar with bilingual NMT, the MNMT model is also trained with the same objective as Eq.\ref{eq:MLE_loss}.

\paragraph{Model Selection Strategy}
The common practice saves a checkpoint at the end of each training epoch, and evaluates its performance on a set of development sets $D^{dev}=\{D^{dev}_1,...,D^{dev}_l,...,D^{dev}_L\}$.
Finally the checkpoint with minimal average dev loss is selected as the \textit{overall best checkpoint}.
This average dev loss could be formalized as:
\begin{align}
\label{eq:dev loss}
    \mathcal{L}^{dev}(\theta, D^{dev})=\frac{1}{L}\sum\limits_{l=1}^{L}\mathcal{L}(D^{dev}_l;\theta).
\end{align}
In our work, we record language-specific dev losses $\mathcal{L}^{dev}_l$ and save the \textit{language-specific best checkpoint} towards each language pair $l$ additionally.

\begin{figure*}[t]
  \centering
    \includegraphics[clip,width=1.9\columnwidth,]{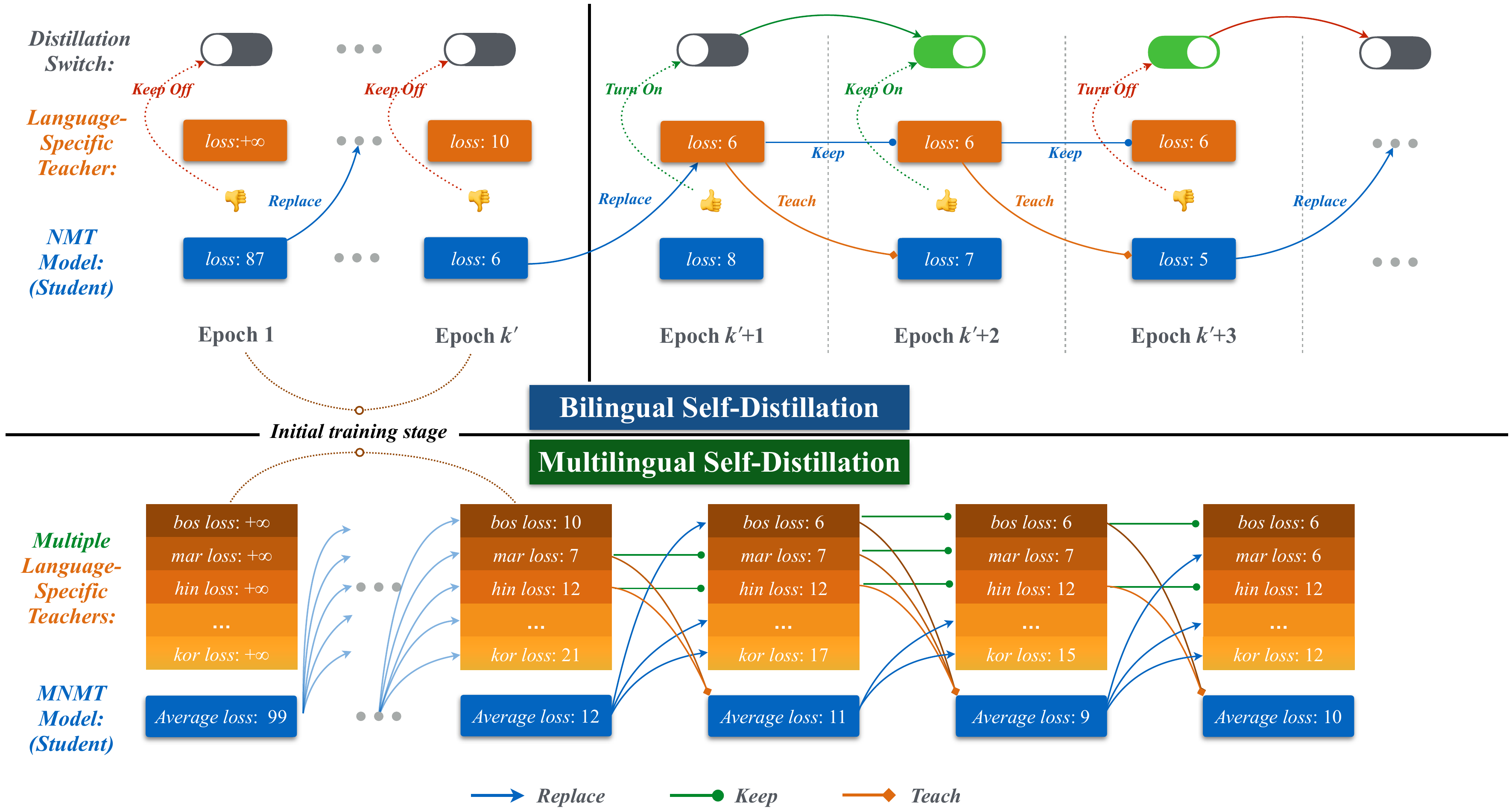}
    \caption{
    The upper part portrays the process of bilingual LSSD. The multilingual LSSD process is shown in the bottom part.
    The blue boxes indicates the student model (NMT Model or multilingual NMT model) and the inside value represents its dev loss in the current training epoch.
    The orange box indicates the teacher model and its loss means its dev loss in the corresponding training epoch.
    The distillation switch, which determines whether to perform distillation in the current epoch, is drawn as an ellipse with grey for off and green for on.}
  \label{fig:LSSD_example}
\end{figure*}

\subsection{Self-Distillation}
Knowledge distillation is an effective model compression technology that distills knowledge from a high-capacity teacher model into the compact student model~\citep{hinton2015distilling}. Self-distillation is an intriguing variation on knowledge distillation with the fundamental difference that self-distillation uses the same network for both the teacher and student model.~\citep{8953762,Zhang_2019_ICCV}.
\citet{8953762} use models in earlier epochs to guide the training of later epochs, which boosts the predictive accuracy in image classification by a large margin.
In this paper, we extend this idea to let the student model learn multiple different targets in the same training epoch and the number of targets is constantly changing in different epochs.

\section{Language-Specific Self-Distillation}
In this section, we first introduce the overall process of LSSD. 
Then, we provide a detailed formalization of LSSD, with special emphasis on the three fine-grained manipulations on knowledge transferring.

\subsection{Overall} \label{subsection:LSSD}

In this subsection, we present an overview of our training strategy for multilingual NMT, as illustrated in Figure \ref{fig:LSSD_example}. 
Specifically, we first take bilingual self-distillation as an example to show the distillation learning process.
Then, we describe the multilingual self-distillation model and how it performs self-distillation towards multiple language pairs.

\paragraph{Bilingual LSSD}
Traditional bilingual neural machine translation builds a Sequence-to-Sequence model for training.
In our bilingual LSSD, we introduce an additional teacher model and a distillation switch.
The teacher model is used to guide the original machine translation model and the switch is designed to decide whether the teacher model is working or not in the current training epoch.
We depict the bilingual LSSD process in the upper half of Figure ~\ref{fig:LSSD_example}.

Normally, at the beginning of training, the loss monotonically decreases, and we refer to this process as ``initial training stage''. 
And $k'$ denotes the number of epochs the initial training stage lasts. It should be noted that the value of $k'$ is not a hyperparameter but is up to the training process.
During these $k'$ epochs, we do not perform distillation (keep the distillation switch off) but replace iteratively the teacher with the better-performance student which has a lower loss.
As we all know, the lower loss, the better. 
Therefore, if the teacher's loss is lower than the student's loss, we turn on the switch, as shown in the $k'$+1 epoch.
In fact, we do not perform distillation due to the switch being closed at the beginning of this epoch.
In the $k'$+2 epoch, since the switch is on, the teacher model distills the student model.
And since the teacher's loss is still lower than the student's loss, the switch remains on.
In the $k'$+3 epoch, the switch also shows turned on, the distillation learning is performed.
But the teacher's loss is higher than the student's loss, we turn off the switch.
At the same time, we replace the teacher model with the current student model to complete the teacher updating.

\paragraph{Multilingual LSSD}

Multilingual LSSD is a complex version of bilingual LSSD, which needs to maintain multiple language-specific teachers and conduct multi-objective distillation learning.
We illustrate the multilingual LSSD process in the bottom half of Figure~\ref{fig:LSSD_example}.
The blue box also means the student model (the current training MNMT model) and the value of loss represents the average loss of the student over multiple language-specific dev sets in the corresponding training epoch.
The different kinds of orange boxes represent different language-specific teacher models, and each loss means different language dev loss in the current training epoch.
For example, the darkest orange box \texttt{bos} refers to the teacher model in \texttt{Bosnian}-to-\texttt{English} translation.

At the beginning of training, multilingual LSSD also has an ``initial training stage'' for each language pair just like the bilingual LSSD, at which time the language-specific teacher doesn't work. Note that the initial training stage of different language pairs may last different epochs.
After the initial training stage, the language-specific teacher model begins working, and the student model can be guided by the teacher.
The different kinds of orange lines mean the language-specific teacher is distilling the student model and the number of working teachers is determined by the language-specific switches, which are consistent with the distillation switch of bilingual LSSD.
The blue line represents the language-specific teacher is replaced by the better-performance student from the last training epoch.
In this paper, multilingual LSSD is the superposition of multiple bilingual LSSDs and they do not affect each other.
We also summarize this process in Algorithm~\ref{alg:algorithm}.

\begin{algorithm}[t]
\label{alg:LSSD}
\small
\caption{Language-Specific Self-Distillation} \label{alg:algorithm}
 \textbf{Input}: 
    \textit{language pairs number} $L$; 
    \textit{training sets} $\{D^{train}_l\}_{l=1}^{L}$; 
    \textit{dev sets} $\{D^{dev}_l\}_{l=1}^{L}$; 
    \textit{max training epochs} $K$; \textit{max training steps of one epoch } $T$;
    \textit{learning rate} $\eta$;\\
 \textbf{Initialize}: 
    initialize MNMT model $\theta$; for $l \in [1, L]$, set $\hat{\theta}_l=\varnothing$, $\Omega_l=off$, $\hat{\mathcal{L}}^{dev}_l=+\infty$.
\begin{algorithmic}[1] 

\For{$k \in [1, K]$} \Comment{For each training epoch}
\For{$t \in [1, T]$} \Comment{Training stage}
\State \textit{Randomly sample a language pair} $l$. 
\State \textit{Sample a mini-batch of sentence pairs} $B^t_l$.
\If { $\Omega_l$ is on}
\State $\mathcal{L} = \mathcal{L}_{NMT}(B^t_l;\theta) + \alpha \mathcal{L}_{LSSD}(B^t_l;\theta,\hat{\theta}_l)$
\Else
\State $\mathcal{L} = \mathcal{L}_{NMT}(B^t_l;\theta)$
\EndIf
\State Update $\theta$: $\theta = \theta - \eta \cdot \nabla_{\theta}\mathcal{L}$
\EndFor

\For{$l \in [1, L]$} \Comment{Validation Stage}
\State $\mathcal{L}^{dev}_l = \mathcal{L}_{NMT}(D^{dev}_l;\theta)$
\If{$\mathcal{L}^{dev}_l < \hat{\mathcal{L}}^{dev}_l$}
\State $\Omega_l=off$, $\hat{\theta}_l =\theta$, $\hat{\mathcal{L}}^{dev}_l=\mathcal{L}^{dev}_l$
\Else
\State $\Omega_l=on$
\EndIf
\EndFor
\EndFor
\end{algorithmic}
\end{algorithm}

\subsection{Formalization of LSSD} 
Formally, we denote the current model by $\theta$ and maintain a set of language-specific best checkpoints $\{\hat{\theta}_l\}_{l=1}^L$. In the validation stage, which is at the end of each epoch, we evaluate the performance of $\theta$ on each language pair. 
For each language pair $l$, if $\theta$ outperforms $\hat{\theta}_l$ in dev set $D^{dev}_l$, $\theta$ replaces $\hat{\theta}_l$. 
To control whether to perform teaching in the current epoch, we define a set of distillation switches $\{\Omega_l\}_{l=1}^L$. 
In the validation stage, if the current model $\theta$ exceeds the language-specific best checkpoint $\hat{\theta}_l$, we turn off the distillation switch, not performing teaching in the next epoch. 
Conversely, when the language-specific best checkpoint wins, we turn on the switch, performing teaching in the next epoch.

When the language-specific distillation switch is on,
the parameters of the current model $\theta$ is updated by optimizing both 
$\mathcal{L}_\text{NMT}$ and $\alpha \cdot \mathcal{L}_\text{LSSD}$.
When the language-specific distillation switch is off, the model $\theta$ is updated only by $\mathcal{L}_\text{NMT}$, \ie{$\alpha  =0$}.
The training loss is calculated as:
\begin{align}
    \label{eq:total loss}
    \mathcal{L} &= \mathcal{L}_\text{NMT} + \alpha \mathcal{L}_\text{LSSD},
\end{align}
where the $\alpha$ is the weight of distillation loss. And $\mathcal{L}_\text{LSSD}$ is computed as the cross-entropy between the output distribution of $\hat{\theta}_l$ and $\theta$, which is formalized as:

\begin{equation}
\begin{aligned}
    \mathcal{L}_\textsc{LSSD} = -\sum\limits_{i \le |y|} \sum\limits_{w \in \mathcal{V}} &P(w|x,y_{<i};\hat{\theta}_l) \\
    &\cdot \log P(w|x,y_{<i};\theta).
\end{aligned}
\end{equation}


\begin{table*}[!t]
\small
\centering
\begin{tabular}{l|cc|cc|cc}

\toprule
\multirow{2}{*}{\textbf{Method}} & \multicolumn{2}{c|}{\textbf{\textsc{Diverse}}} & \multicolumn{2}{c|}{\textbf{\textsc{Related}}} & \multicolumn{2}{c}{\textbf{\textsc{WMT}}} \\
& \textbf{M2O} & \textbf{O2M} & \textbf{M2O} & \textbf{O2M} & \textbf{M2O} & \textbf{O2M} \\

\midrule
\midrule
\multicolumn{7}{c}{\textit{Baselines}} \\
\midrule

\textsc{Multilingual} & 29.00 & 22.85 & 27.83 & 21.85 & 20.15 & 19.07\\
\textsc{Multi-Distill}~\cite{tan-etal-2018-multilingual} &  29.52 & 22.31 & 26.60 & 21.70 & 20.18 & 18.57 \\

\midrule
\midrule
\multicolumn{7}{c}{\textit{Previous Works}} \\
\midrule

MultiDDS-S~\cite{wang-etal-2020-balancing}$^*$
& 27.00 & 18.24 & 25.52 & 17.32 & -- & -- \\
MultiUAT~\cite{wu-etal-2021-uncertainty}$^*$
& 27.83 & 19.76 & 26.39 & 18.64 & -- & -- \\
CCL-M~\cite{zhang-etal-2021-competence-based}$^*$
& 28.34 & 19.53 & 26.73 & 18.89 & -- & -- \\
$\chi$-IBR~\cite{zhou-etal-2021-distributionally}$^*$
& 29.74 & 23.44 & 28.71 & 22.21 & -- & -- \\

\midrule
\midrule
\multicolumn{7}{c}{\textit{Our Proposed Approaches}} \\
\midrule

LSSD-Whole  & 30.57$^\dagger$ & \textbf{23.55}$^\dagger$ & 29.28$^\dagger$ & 22.20$^\dagger$ & 21.05$^\dagger$ & \textbf{19.76}$^\dagger$\\
LSSD-Selective & 30.24$^\dagger$ & 23.16 & 28.65$^\dagger$ & 22.15$^\dagger$ & \textbf{21.17}$^\dagger$ & 19.32$^\dagger$ \\
LSSD-Adaptive & \textbf{30.77}$^\dagger$ & 23.39 & \textbf{29.40}$^\dagger$ & \textbf{22.27}$^\dagger$ & 20.96$^\dagger$ & 19.48$^\dagger$\\
\bottomrule
\end{tabular}
\caption{BLEU scores on TED-8-Diverse (\textsc{Diverse}) and TED-8-Related (\textsc{Related}) and WMT datasets.
Bold indicates the highest BLEU value on each setting. 
`*' represents results taken from original papers.
 ``M2O'' means Many-to-One translation. ``O2M'' means One-to-Many translation. 
 `$\dagger$' means significantly better than \textsc{Multilingual} with t-test $p<0.01$.}
\label{tab:main results}
\end{table*}

\subsection{Sample-level Manipulations for LSSD} \label{subsection:IASD}
To prevent the potential negative transferring on some translation samples where the teacher underperforms the student, we devise three sample-level manipulations for LSSD: \textsc{LSSD-Whole}, \textsc{LSSD-Selective} and  \textsc{LSSD-Adaptive}. All of them could be generalized as rescaling the distillation loss with a sample-level weight:
\begin{align}
    \mathcal{L}_\textsc{LSSD} = \mathcal{L}_\textsc{LSSD} \times \mathcal{G},
\end{align}
where $\mathcal{G}$ is the sample-level weight which is determined by the performance difference between teacher and student. And different operations correspond different implementation of $\mathcal{G}$.

\paragraph{\textsc{LSSD-Whole}} In this manner, we equally execute distillation with the same sample-level weight on all data samples. 
In practice, it is equivalent to setting $\mathcal{G}=1$.
\textsc{LSSD-Whole} can be viewed as the base version of LSSD.

\paragraph{\textsc{LSSD-Selective}} To avert the negative impact of performing distilltaion on data samples where the teacher model errs on, we explore to only distilling samples where the teacher performs better than the student. And this is implemented as:
\begin{align}
    \mathcal{G}(x,y,\theta, \hat{\theta}_l)=\left\{
\begin{aligned}
0, &  &P(y|x;\hat{\theta}_l) < P(y|x;\theta)& \\
1, &  &P(y|x;\hat{\theta}_l) \ge P(y|x;\theta)&, \\
\end{aligned}
\right.
\end{align}
where $P(y|x;\theta)$ is the likelihood probability that model $\theta$ outputs with respect to the target sentence $y$ given the input $x$. And we compute the sentence probability by averaging probabilities over all tokens.

\paragraph{\textsc{LSSD-Adaptive}} Since a one-size-fits-all rule prohibiting distillation on a subset of samples could limit flexibility, we design \textsc{LSSD-Adaptive} changing the distillation weight according to the teacher-student performance ratio, which is implemented as:
\begin{align}
\label{eq:teacher-student performance ratio}
\mathcal{G}(x,y,\theta, \hat{\theta}_l) = min(\frac{P(y|x;\hat{\theta}_l)}{P(y|x;\theta)}, \sigma),
\end{align}
where $\sigma$ is a hyperparameter to truncate the $\mathcal{G}$ when higher than $\sigma$, value of which is set to 2 empirically\footnote{We search the optimal $\sigma$ in $\{1.0, 1,5, 2.0\}$}.

\begin{table*}[t]
\renewcommand\tabcolsep{4.0pt}
\small
    \centering
    \begin{tabular}{c | c | c | cccccccc |c}
    \toprule
        \textbf{Dataset} & \textbf{Setting} & \textbf{Method} & \textbf{bos} & \textbf{mar} & \textbf{hin} & \textbf{mkd} & \textbf{ell} & \textbf{bul} & \textbf{fra}   & \textbf{kor} & \textbf{Avg.} \\
    \midrule
        \multirow{4}{*}{\rotatebox{90}{\textsc{Diverse\ \ }}} &
        \multirow{2}{*}{M2O} & \textsc{Multi-Distill} & -0.76 & -0.39 & -0.86 & +0.58 & +1.65 & +1.23 & +1.30 & +1.36 & +0.52 \\
        & & LSSD & \textbf{+1.98} & \textbf{+1.00} & \textbf{+1.30} & \textbf{+2.17} & \textbf{+2.24} & \textbf{+1.81} & \textbf{+1.54} & \textbf{+2.10} & \textbf{+1.77} \\
        \cmidrule{2-12}
        & \multirow{2}{*}{O2M} & \textsc{Multi-Distill} & -3.44 & -0.87 & -1.26 & +0.41 & +0.49 & +0.35 & \textbf{+0.06} & -0.04  & -0.54 \\
        & & LSSD & \textbf{+0.93} & \textbf{+0.37} & \textbf{+0.52} & \textbf{+2.37} & \textbf{+0.75} & \textbf{+0.39} & +0.04 & \textbf{+0.20} & \textbf{+0.70} \\
    \midrule
    \midrule
    
        & & & \textbf{aze} & \textbf{bel}	& \textbf{glg} & \textbf{slk} & \textbf{tur} & \textbf{rus} & \textbf{por} & \textbf{ces}	& \textbf{Avg.}\\
    \midrule
        \multirow{4}{*}{\rotatebox{90}{\textsc{Related\ \ }}} &
        \multirow{2}{*}{M2O} & \textsc{Multi-Distill} & -4.31 & -6.38 & -4.87 & +0.09 & \textbf{+1.52} & \textbf{+1.16} & \textbf{+1.62} & +1.33 & -1.23 \\
        & & LSSD & \textbf{+1.53} & \textbf{+2.35} & \textbf{+1.70} & \textbf{+1.47} & +1.18 & +1.04 & +1.43 & \textbf{+1.82} & \textbf{+1.57}\\
        \cmidrule{2-12}
        & \multirow{2}{*}{O2M} & \textsc{Multi-Distill} & -0.32 & -0.64 & +0.14 & +0.27 & -0.20 & -0.17 & +0.09 & -0.31 & -0.15 \\
        & & LSSD & \textbf{+0.15} & \textbf{+0.12} & \textbf{+0.50} & \textbf{+0.32} & \textbf{+0.68} & \textbf{+0.20} & \textbf{+0.54} & \textbf{+0.32} & \textbf{+0.35} \\
    \midrule
    \midrule
    
        & & & \textbf{tr} & \textbf{ro} & \textbf{et} & \textbf{zh} & \textbf{de} & \textbf{fr} & -- & -- & \textbf{Avg.} \\
    \midrule
        \multirow{4}{*}{\rotatebox{90}{\textsc{WMT\ \ }}} & \multirow{2}{*}{M2O} & \textsc{Multi-Distill} & -0.84 & -1.40 & +0.10 & \textbf{+0.78} & \textbf{+0.71} & \textbf{+0.82} & -- & -- & +0.03 \\
        & & LSSD & \textbf{+0.79} & \textbf{+1.3} & \textbf{+1.29} & +0.32 & +0.5 & +0.68 & -- & -- & \textbf{+0.81} \\
        \cmidrule{2-12}
        & \multirow{2}{*}{O2M} & \textsc{Multi-Distill} & -2.42 & -2.26 & -0.69 & +0.58 & \textbf{+0.87} & \textbf{+0.92} & -- & -- & -0.50 \\
        & & LSSD & \textbf{+0.81} & \textbf{+0.53} & \textbf{+0.48} & \textbf{+1.00} & +0.68 & +0.66& -- & -- & \textbf{+0.69} \\
    \bottomrule
    \end{tabular}
    \caption{BLEU score improvements of \textsc{Multi-Distill} and our LSSD over the \textsc{Multilingual} baseline.
    For clarity, we take \textsc{LSSD-Adaptive} and \textsc{LSSD-Whole} as the representatives of LSSD in M2O and O2M translation respectively.
    Bold indicates the best performance. 
    Languages are ordered increasingly by data size from left to right.}
    \label{tab:results per languages}
    \vspace{-3mm}
\end{table*}


\section{Experiments}

\subsection{Settings}

\paragraph{Datasets}
We conduct experiments on three datasets: the widely-used TED-8-Diverse and TED-8-Related~\citep{wang-etal-2020-balancing}, and a relative large-scale WMT dataset. 
The TED-8-Diverse contains 4 low-resource languages (\texttt{bos}, \texttt{mar}, \texttt{hin}, \texttt{mkd}) and 4 high-resource languages (\texttt{ell}, \texttt{bul}, \texttt{fra}, \texttt{kor}) to English. 
The TED-8-Related contains 4 low-resource languages (\texttt{aze}, \texttt{bel}, \texttt{glg}, \texttt{slk}) and 4 related high-resource language (\texttt{tur}, \texttt{rus}, \texttt{por}, \texttt{ces}) to English. 
Both of these two datasets have around 570K sentence pairs.
We detail the data statistics and the interpretation of language codes in Appendix~\ref{section:data statistics}.

For the WMT dataset, we consider 3 low-resource languages (\texttt{et}, \texttt{ro}, \texttt{tr}) and 3 high-resource languages (\texttt{fr}, \texttt{de}, \texttt{zh}) to English.
Totally around 5M training sentences are sampled from the parallel corpus provided by WMT14, WMT16, WMT17, and WMT18. And we use the corresponding dev and test sets for validation and evaluation. The detailed data statistics are also placed in Appendix~\ref{section:data statistics}. Compared to TED-8-Diverse and TED-8-related, the size of the WMT dataset is larger and distributed more unevenly over various languages.

For each dataset, we experiment in two multilingual translation scenarios: 1) \textsc{Many-to-One}~(M2O): translating multiple languages to English in this work; 2) \textsc{One-to-Many}~(O2M): translating English to different languages. 

\paragraph{Hyperparameters}
We verify the effectiveness of LSSD on the Transformer~\cite{vaswani-etal-2017-attention} as implemented in fairseq~\cite{ott-etal-2019-fairseq} with 6 layers and 8 attention heads. And we use the same hyperparameters with the previous SOTA~\citep{zhou-etal-2021-distributionally} to obtain a strong baseline. The only difference with~\citet{zhou-etal-2021-distributionally} is that we train all models for 300 epochs which is less than theirs\footnote{We also tried for more epochs, while no performance gains happened}.
For the TED-8-Diverse and TED-8-Related, we follow previous works \citet{wang-etal-2020-balancing,zhang-etal-2021-competence-based} to preprocess both datasets using sentencepiece~\citep{kudo-richardson-2018-sentencepiece} with a vocabulary size of 8\textit{K} for each language. For the WMT dataset, we preprocess data using sentencepiece with a vocabulary size of 64\textit{K} for all languages.
The complete set of hyperparameters can be found in Appendix~\ref{subsection:hyperparameters}.
All models are trained on 8 Tesla V100 GPUs.
And the performance is evaluated with BLEU score using sacreBLEU~\cite{papineni-etal-2002-bleu,post-2018-call}.
We set distillation weight $\alpha$ to 2.0 in M2O and 0.6 in O2M respectively (see section~\ref{subsection: Effect of Distillation Weight} for analysis on these choices).

\paragraph{Baselines}
We compare our LSSD with: 
1) the standard-trained multilingual NMT model (\ie \textsc{Multilingual})~\citep{johnson-etal-2017-googles}; 
2) \textsc{Multi-Distill}~\citep{tan-etal-2018-multilingual}, which is also a distillation-based strategy that guides the multilingual model in each translation direction utilizing bilingual models. To re-implement \textsc{Multi-Distill}, we first train bilingual models of each language pair on all three datasets\footnote{we report performance of bilingual models in Appendix~\ref{subsection: bilingual performance}}. 
Follow previous works~\citep{tan-etal-2018-multilingual,zhang-etal-2021-competence-based}, we train bilingual models using the same model configuration and hyper-parameters with multilingual models.
For all baselines and our LSSD, the same model configuration and hyper-parameters are applied.

\begin{figure*}[th]
  \centering
    \includegraphics[clip,width=2.0\columnwidth,]{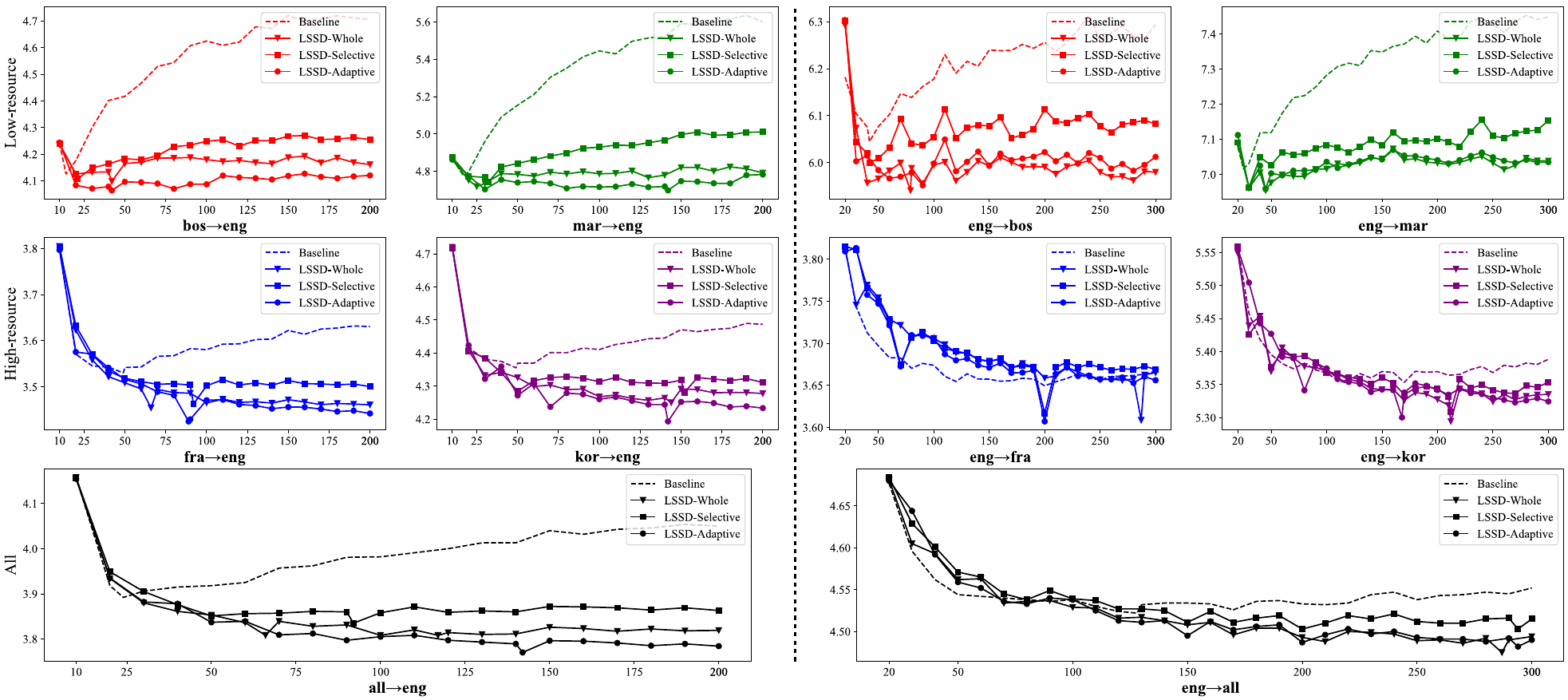}
    \caption{ Loss curves of the \textsc{Multilingual} baseline and LSSD in M2O (left half part) and O2M (right half part) settings on the TED-8-Diverse dataset. 
     The x-axis and y-axis indicate training epochs and dev losses respectively. 
     Due to space limitation, we only display 4 language pairs (2 low-resource + 2 high-resource) in each setting.}
  \label{fig:loss curve Diverse}
\end{figure*}

\subsection{Main Results}
\paragraph{Overall results} 
We summarize main results into Table~\ref{tab:main results}. As we can see, 
1) on all three datasets, our LSSD significantly outperforms the baselines under M2O and O2M settings, demonstrating the effectiveness of our approach;
2) compared with previous works, LSSD achieves higher BLEU scores on the TED-8-Diverse and TED-8-Related datasets, which indicates the superiority of our method;
3) in the comparison among the three variants of LSSD, \textsc{LSSD-Adaptive} excels in M2O and \textsc{LSSD-Whole} performs best in O2M overall. To better understand this phenomenon, we analyzed the teacher-student performance ratio (detailed in Equation~\ref{eq:teacher-student performance ratio}) in M2O and O2M respectively, and discovered that this ratio varies more significantly in the challenging O2M translation than in the M2O (the variance is 0.097 in O2M and is 0.039 in M2O), which may incur an unstable training for LSSD-Adaptive.

\paragraph{Results on each language}
Looking closer at results per languages for the \textsc{Multi-Distill} and our LSSD,
we calculate the difference between \textsc{Multilingual} and \textsc{Multi-Distill} or LSSD separately in terms of BLEU, which is shown in Table~\ref{tab:results per languages}.
Firstly, on all datasets and settings, \textbf{LSSD consistently outperforms the \textsc{Multilingual} baseline across all language pairs}. The improvement is up to 2.37 (\texttt{eng} $\rightarrow$\texttt{mkd}).
Secondly, Multi-Distill struggles in low-resource directions across all three datasets, which is not unexpected considering that the training data is too scarce to train trustworthy bilingual teacher models. 
However, LSSD breaks this limitation, obtaining stable improvements on \textbf{both} low-resource and high-resource language pairs, by employing multilingual NMT models that own more balanced performance as teacher models.
\begin{table}[t]
\small
\renewcommand\tabcolsep{2.0pt}
\centering
\begin{tabular}{l|cc|cc|cc}

\toprule
\multirow{2}{*}{\textbf{Method}} & \multicolumn{2}{c|}{\textsc{\textbf{Diverse}}} & \multicolumn{2}{c|}{\textsc{\textbf{Related}}} & \multicolumn{2}{c}{\textsc{\textbf{WMT}}} \\
& M2O & O2M & M2O & O2M & M2O & O2M \\
\midrule

\textsc{Multilingual} & 0.27 & 0.64 & 0.57 & 0.51 & 0.75 & 1.09\\

\textsc{LSSD-Whole} & 0.21 & \textbf{0.15} & 0.32 & 0.30 & 0.25 & \textbf{0.32} \\

\textsc{LSSD-Selective} & 0.24 & 0.37 & 0.35 & 0.29 & 0.21 & 0.36 \\

\textsc{LSSD-Adaptive} & \textbf{0.19} & 0.35 & \textbf{0.16} & \textbf{0.26} & \textbf{0.14} & 0.44 \\

\bottomrule
\end{tabular}
\caption{Performance deficit of the \textsc{Multilingual} baseline and our LSSD. Bold indicates the lowest value under each setting. Compared to the baseline, LSSD reduces this deficit by 57\% on average.}
\label{tab:distance to upper bound of all method in all datsets and settings.}
\end{table}

\section{Analysis}

\subsection{Convergence Inconsistency}
\label{subsection:Convergence Inconsistency}
In this work, we propose to formalize the convergence inconsistency as the loss gap between the overall best checkpoint and language-specific best checkpoints, which is referred to as ``{\textbf{Performance Deficit}})''. 
Concretely, for each language pair $l$, we accumulate the dev loss differences between the overall best checkpoint $\theta$ and the corresponding language-specific best checkpoint $\hat{\theta}_l$.
We give a formal definition as:
\begin{align}
    \text{DUB}(\theta, \{\hat{\theta_l}\}^L_{l=1}) \!=\! \sum_{l=1}^{L} (\mathcal{L}(\theta, D_l^{dev})
    \!-\! \mathcal{L}(\hat{\theta}_l, D_l^{dev})).
\end{align}
Note that the performance deficit is a non-negative value because $\mathcal{L}(\theta, D_l^{dev})$ is always greater than or equal to $\mathcal{L}(\hat{\theta}_l, D_l^{dev})$.
We list the performance deficit of the \textsc{Multilingual} baseline and our LSSD in Table~\ref{tab:distance to upper bound of all method in all datsets and settings.}. 
As observed, the performance deficit of LSSD is significantly lower than the baseline (decreased 57\% on average), which proves the efficacy of LSSD in remedying convergence inconsistency.

\subsection{Multiple Teachers vs. Single Teacher}
\begin{table}[t]
\small
\renewcommand\tabcolsep{2.5pt}
\centering
\begin{tabular}{l|cc|cc|cc}

\toprule
\multirow{2}{*}{\textbf{Method}} & \multicolumn{2}{c|}{\textsc{Diverse}} & \multicolumn{2}{c|}{\textsc{Related}} & \multicolumn{2}{c}{\textsc{WMT}} \\

& \textbf{M2O} & \textbf{O2M} & \textbf{M2O} & \textbf{O2M} & \textbf{M2O} & \textbf{O2M} \\
\midrule

LSSD &\textbf{+1.57} & \textbf{+0.70} & \textbf{+1.45} & \textbf{+0.35} & \textbf{+0.90} & \textbf{+0.69} \\

STSD & +0.89 & +0.00 & +0.86 & -0.14 & +0.18 & -0.20 \\
\bottomrule
\end{tabular}
\caption{BLEU score improvements of LSSD (Language-Specific Self-Distillation) and STSD (Single-Teacher Self-Distillation) over the \textsc{Multilingual} baseline on all three datasets.}
\label{tab:multiple teachers vs. single teacher.}
\end{table}
To provide light on the necessity of using language-specific best checkpoints as teacher models, we conduct ablation study in Table~\ref{tab:multiple teachers vs. single teacher.}. The STSD (Single Teacher Self-Distillation) uses the overall best checkpoint instead of language-specific best checkpoints to guide the training of multilingual NMT models. 
To make a fair comparison, we report the results of \textsc{LSSD-Whole}.
As indicated, the gains from STSD over the baseline only account for 20\% to 60\% of LSSD's gains in the M2O translation.
Even more, STSD fails to enhance the O2M translation. 
These results show that the multilingual model gains a great deal from tailored language-specific teachers indeed.

\subsection{Comparison of Loss Curves}
\label{subsection:Effectiveness on Alleviating Over-fitting}
To better comprehend how each approach affects the model training process, we analyze different methods from the perspective of loss curves, as illustrated in Figure~\ref{fig:loss curve Diverse}. 
Firstly, comparing the loss curves of the \textsc{Multilingual} baseline with LSSD (solid lines vs. dotted lines), it is clear that the baseline suffers from serious over-fitting in low-resources (\eg{\texttt{bos} $\leftrightarrow$ \texttt{eng}}). By letting the model \textit{recall} previous checkpoints, LSSD mitigates the over-fitting, which delays the convergence to lengthen the training time for high-resource languages.
Secondly, comparing the loss curves of baselines in M2O and O2M (dotted lines in left vs. right half section), it is observed that M2O suffers from more serious over-fitting than O2M. 
This explains why M2O benefits more from LSSD than O2M.
Lastly, contrasting the three modes of \textsc{LSSD}, \textsc{LSSD-Adaptive} performs better in M2O and achieves comparable with \textsc{LSSD-Whole} in O2M in terms of dev loss.

\subsection{Effect of Distillation Weight $\alpha$}
\label{subsection: Effect of Distillation Weight}
\begin{figure}[t]
  \centering
    \includegraphics[clip,width=1.0\columnwidth,]{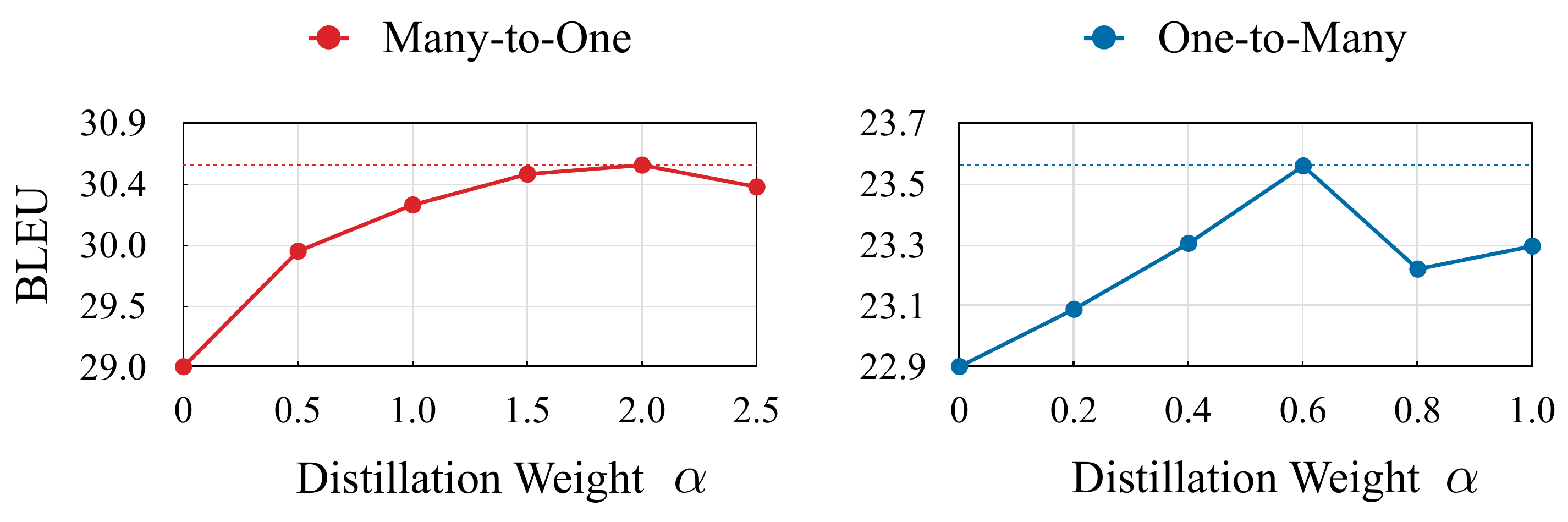}
    \caption{Effect of different distillation weights on LSSD. For clarity, we take \textsc{LSSD-Whole} as the representative of LSSD.}
  \label{fig:Effect on weight}
\end{figure}

As Equation~\ref{eq:total loss} shows, LSSD trains multilingual NMT models with NMT loss and the $\alpha$-weighted distillation loss jointly. We demonstrate the effect of different $\alpha$ on LSSD in Figure~\ref{fig:Effect on weight}. 
As we can see, the optimal weight for O2M ($\alpha=0.6$) is smaller than which for M2O ($\alpha=2.0$).
We conjecture this is due to the fact that O2M converges later than M2O (see Section~\ref{subsection:Effectiveness on Alleviating Over-fitting}), meaning that O2M models might learn from immature teachers for more epochs. Consequently, a lower distillation strength is more suited to eliminating this danger.

\section{Related Works}
\subsection{Advances in Multilingual NMT}
Recently, multilingual NMT mainly focuses on: 1) designing effective parameters sharing strategy \cite{zhang-etal-2021-share,zhu-etal-2021-counter-interference,xie-etal-2021-importance,lin-etal-2021-learning}; 2) obtaining language-agnostic representations ~\cite{zhu-etal-2020-language,pan-etal-2021-contrastive}; 3) incorporating pre-training models ~\cite{siddhant-etal-2020-leveraging,wang-etal-2020-multi}; 4) resolving the data imbalance among diverse languages~\cite{wang-etal-2020-balancing,wu-etal-2021-uncertainty,zhang-etal-2021-competence-based,zhou-etal-2021-distributionally}.
Different from them, LSSD is designed for alleviating the convergence inconsistency, which is ignored by existing works.

\subsection{Knowledge Distillation in NMT}
To the best of our knowledge, \newcite{kim-rush-2016-sequence} first apply knowledge distillation in bilingual NMT and propose a sequence-level distillation.
\newcite{wei-etal-2019-online} propose to avoid over-fitting by guiding the training process with best checkpoints. 

In multilingual NMT, \newcite{tan-etal-2018-multilingual} use knowledge distillation to close the gap between the multilingual NMT model and bilingual models. 
However, their work is based on the hypothesis that there are sufficient training data for each language pair to prepare a promising bilingual teacher. 
In fact, most languages face the resource-scarcity problem. 
In our work, LSSD's teacher is the multilingual model, which has a better performance for low resource language translation via transfer learning.
And we further develop three sample-level operations for LSSD via weighing teacher and student performance.

\section{Conclusion}
In this work, we propose a novel training strategy \textbf{L}anguage-\textbf{S}pecific \textbf{S}elf-\textbf{D}istillation (LSSD) to remedy the convergence inconsistency in multilingual neural machine translation. Moreover, we devise three sample-level manipulations for LSSD. 
Experimental results on three datasets demonstrate that LSSD achieves SOTA performance. 
Through analysis experiments, we also find that: 1) LSSD significantly mitigates the convergence inconsistency (decreased 57\% on average), which is quantified by performance deficit;
2) both low-resource and high-resource languages benefit from LSSD.
\section{Limitations}
Same with other distillation-based works~\citep{hinton2015distilling,tan-etal-2018-multilingual}, LSSD takes some extra training overhead. Taking the training in TED-8-Diverse M2O as an example, the \textsc{Baseline} and \textsc{LSSD-Whole} spend 5.7 hours and 7.6 hours respectively. 
However, it is worth noting that our method doesn't affect the inference speed of the model.

\section*{Acknowledgements}
Xiaocheng Feng is the corresponding author of this work. We thank the anonymous reviewers for their insightful comments. This work was supported by the National Key R\&D Program of China via grant 2020AAA0106502, National Natural Science Foundation of China (NSFC) via grant 62276078 and the Major Key Project of PCL, PCL2022D01.

\bibliography{anthology,custom}

\begin{thebibliography}{50}
\expandafter\ifx\csname natexlab\endcsname\relax\def\natexlab#1{#1}\fi

\bibitem[{Aharoni et~al.(2019)Aharoni, Johnson, and
  Firat}]{aharoni-etal-2019-massively}
Roee Aharoni, Melvin Johnson, and Orhan Firat. 2019.
\newblock \href {https://doi.org/10.18653/v1/N19-1388} {Massively multilingual
  neural machine translation}.
\newblock In \emph{Proceedings of the 2019 Conference of the North {A}merican
  Chapter of the Association for Computational Linguistics: Human Language
  Technologies, Volume 1 (Long and Short Papers)}, pages 3874--3884,
  Minneapolis, Minnesota. Association for Computational Linguistics.

\bibitem[{Arivazhagan et~al.(2019)Arivazhagan, Bapna, Firat, Lepikhin, Johnson,
  Krikun, Chen, Cao, Foster, Cherry, Macherey, Chen, and
  Wu}]{arivazhagan-etal-2019-massively}
Naveen Arivazhagan, Ankur Bapna, Orhan Firat, Dmitry Lepikhin, Melvin Johnson,
  Maxim Krikun, Mia~Xu Chen, Yuan Cao, George~F. Foster, Colin Cherry, Wolfgang
  Macherey, Zhifeng Chen, and Yonghui Wu. 2019.
\newblock \href {http://arxiv.org/abs/1907.05019} {Massively multilingual
  neural machine translation in the wild: Findings and challenges}.
\newblock \emph{CoRR}, abs/1907.05019.

\bibitem[{Bahdanau et~al.(2015)Bahdanau, Cho, and
  Bengio}]{bahdanau-etal-2015-neural}
Dzmitry Bahdanau, Kyunghyun Cho, and Yoshua Bengio. 2015.
\newblock Neural machine translation by jointly learning to align and
  translate.
\newblock In \emph{ICLR}.

\bibitem[{Cho et~al.(2014)Cho, van Merri{\"e}nboer, Gulcehre, Bahdanau,
  Bougares, Schwenk, and Bengio}]{cho-etal-2014-learning}
Kyunghyun Cho, Bart van Merri{\"e}nboer, Caglar Gulcehre, Dzmitry Bahdanau,
  Fethi Bougares, Holger Schwenk, and Yoshua Bengio. 2014.
\newblock \href {https://doi.org/10.3115/v1/D14-1179} {Learning phrase
  representations using {RNN} encoder{--}decoder for statistical machine
  translation}.
\newblock In \emph{Proceedings of the 2014 Conference on Empirical Methods in
  Natural Language Processing ({EMNLP})}, pages 1724--1734, Doha, Qatar.
  Association for Computational Linguistics.

\bibitem[{Fan et~al.(2021)Fan, Bhosale, Schwenk, Ma, El-Kishky, Goyal, Baines,
  Celebi, Wenzek, Chaudhary, Goyal, Birch, Liptchinsky, Edunov, Auli, and
  Joulin}]{fan-etal-2021-beyond}
Angela Fan, Shruti Bhosale, Holger Schwenk, Zhiyi Ma, Ahmed El-Kishky,
  Siddharth Goyal, Mandeep Baines, Onur Celebi, Guillaume Wenzek, Vishrav
  Chaudhary, Naman Goyal, Tom Birch, Vitaliy Liptchinsky, Sergey Edunov,
  Michael Auli, and Armand Joulin. 2021.
\newblock \href {http://jmlr.org/papers/v22/20-1307.html} {Beyond
  english-centric multilingual machine translation}.
\newblock \emph{Journal of Machine Learning Research}, 22(107):1--48.

\bibitem[{Firat et~al.(2016)Firat, Cho, and Bengio}]{firat-etal-2016-multi}
Orhan Firat, Kyunghyun Cho, and Yoshua Bengio. 2016.
\newblock \href {https://doi.org/10.18653/v1/N16-1101} {Multi-way, multilingual
  neural machine translation with a shared attention mechanism}.
\newblock In \emph{Proceedings of the 2016 Conference of the North {A}merican
  Chapter of the Association for Computational Linguistics: Human Language
  Technologies}, pages 866--875, San Diego, California. Association for
  Computational Linguistics.

\bibitem[{Gehring et~al.(2017)Gehring, Auli, Grangier, and
  Dauphin}]{gehring-etal-2017-convolutional}
Jonas Gehring, Michael Auli, David Grangier, and Yann Dauphin. 2017.
\newblock \href {https://doi.org/10.18653/v1/P17-1012} {A convolutional encoder
  model for neural machine translation}.
\newblock In \emph{Proceedings of the 55th Annual Meeting of the Association
  for Computational Linguistics (Volume 1: Long Papers)}, pages 123--135,
  Vancouver, Canada. Association for Computational Linguistics.

\bibitem[{Geng et~al.(2018)Geng, Feng, Qin, and Liu}]{geng-etal-2018-adaptive}
Xinwei Geng, Xiaocheng Feng, Bing Qin, and Ting Liu. 2018.
\newblock \href {https://doi.org/10.18653/v1/D18-1048} {Adaptive multi-pass
  decoder for neural machine translation}.
\newblock In \emph{Proceedings of the 2018 Conference on Empirical Methods in
  Natural Language Processing}, pages 523--532, Brussels, Belgium. Association
  for Computational Linguistics.

\bibitem[{Gu et~al.(2019)Gu, Wang, Cho, and Li}]{gu-etal-2019-improved}
Jiatao Gu, Yong Wang, Kyunghyun Cho, and Victor~O.K. Li. 2019.
\newblock \href {https://doi.org/10.18653/v1/P19-1121} {Improved zero-shot
  neural machine translation via ignoring spurious correlations}.
\newblock In \emph{Proceedings of the 57th Annual Meeting of the Association
  for Computational Linguistics}, pages 1258--1268, Florence, Italy.
  Association for Computational Linguistics.

\bibitem[{Ha et~al.(2016)Ha, Niehues, and Waibel}]{ha-etal-2016-toward}
Thanh-Le Ha, Jan Niehues, and Alex Waibel. 2016.
\newblock \href {https://aclanthology.org/2016.iwslt-1.6} {Toward multilingual
  neural machine translation with universal encoder and decoder}.
\newblock In \emph{Proceedings of the 13th International Conference on Spoken
  Language Translation}, Seattle, Washington D.C. International Workshop on
  Spoken Language Translation.

\bibitem[{Hassan et~al.(2018)Hassan, Aue, Chen, Chowdhary, Clark, Federmann,
  Huang, Junczys{-}Dowmunt, Lewis, Li, Liu, Liu, Luo, Menezes, Qin, Seide, Tan,
  Tian, Wu, Wu, Xia, Zhang, Zhang, and Zhou}]{hassan-etal-2018-achieving}
Hany Hassan, Anthony Aue, Chang Chen, Vishal Chowdhary, Jonathan Clark,
  Christian Federmann, Xuedong Huang, Marcin Junczys{-}Dowmunt, William Lewis,
  Mu~Li, Shujie Liu, Tie{-}Yan Liu, Renqian Luo, Arul Menezes, Tao Qin, Frank
  Seide, Xu~Tan, Fei Tian, Lijun Wu, Shuangzhi Wu, Yingce Xia, Dongdong Zhang,
  Zhirui Zhang, and Ming Zhou. 2018.
\newblock \href {http://arxiv.org/abs/1803.05567} {Achieving human parity on
  automatic chinese to english news translation}.
\newblock \emph{CoRR}, abs/1803.05567.

\bibitem[{Hinton et~al.(2015)Hinton, Vinyals, and Dean}]{hinton2015distilling}
Geoffrey Hinton, Oriol Vinyals, and Jeff Dean. 2015.
\newblock \href {http://arxiv.org/abs/1503.02531} {Distilling the knowledge in
  a neural network}.

\bibitem[{Johnson et~al.(2017)Johnson, Schuster, Le, Krikun, Wu, Chen, Thorat,
  Vi{\'e}gas, Wattenberg, Corrado, Hughes, and
  Dean}]{johnson-etal-2017-googles}
Melvin Johnson, Mike Schuster, Quoc~V. Le, Maxim Krikun, Yonghui Wu, Zhifeng
  Chen, Nikhil Thorat, Fernanda Vi{\'e}gas, Martin Wattenberg, Greg Corrado,
  Macduff Hughes, and Jeffrey Dean. 2017.
\newblock \href {https://doi.org/10.1162/tacl_a_00065} {{G}oogle{'}s
  multilingual neural machine translation system: Enabling zero-shot
  translation}.
\newblock \emph{Transactions of the Association for Computational Linguistics},
  5:339--351.

\bibitem[{Kalchbrenner and Blunsom(2013)}]{kalchbrenner-blunsom-2013-recurrent}
Nal Kalchbrenner and Phil Blunsom. 2013.
\newblock \href {https://aclanthology.org/D13-1176} {Recurrent continuous
  translation models}.
\newblock In \emph{Proceedings of the 2013 Conference on Empirical Methods in
  Natural Language Processing}, pages 1700--1709, Seattle, Washington, USA.
  Association for Computational Linguistics.

\bibitem[{Kim and Rush(2016)}]{kim-rush-2016-sequence}
Yoon Kim and Alexander~M. Rush. 2016.
\newblock \href {https://doi.org/10.18653/v1/D16-1139} {Sequence-level
  knowledge distillation}.
\newblock In \emph{Proceedings of the 2016 Conference on Empirical Methods in
  Natural Language Processing}, pages 1317--1327, Austin, Texas. Association
  for Computational Linguistics.

\bibitem[{Kingma and Ba(2015)}]{DBLP:journals/corr/KingmaB14}
Diederik~P. Kingma and Jimmy Ba. 2015.
\newblock Adam: {A} method for stochastic optimization.
\newblock In \emph{ICLR}.

\bibitem[{Kudo and Richardson(2018)}]{kudo-richardson-2018-sentencepiece}
Taku Kudo and John Richardson. 2018.
\newblock \href {https://doi.org/10.18653/v1/D18-2012} {{S}entence{P}iece: A
  simple and language independent subword tokenizer and detokenizer for neural
  text processing}.
\newblock In \emph{Proceedings of the 2018 Conference on Empirical Methods in
  Natural Language Processing: System Demonstrations}, pages 66--71, Brussels,
  Belgium. Association for Computational Linguistics.

\bibitem[{Lin et~al.(2021)Lin, Wu, Wang, and Li}]{lin-etal-2021-learning}
Zehui Lin, Liwei Wu, Mingxuan Wang, and Lei Li. 2021.
\newblock \href {https://doi.org/10.18653/v1/2021.acl-long.25} {Learning
  language specific sub-network for multilingual machine translation}.
\newblock In \emph{Proceedings of the 59th Annual Meeting of the Association
  for Computational Linguistics and the 11th International Joint Conference on
  Natural Language Processing (Volume 1: Long Papers)}, pages 293--305, Online.
  Association for Computational Linguistics.

\bibitem[{Liu et~al.(2016)Liu, Utiyama, Finch, and
  Sumita}]{liu-etal-2016-neural}
Lemao Liu, Masao Utiyama, Andrew Finch, and Eiichiro Sumita. 2016.
\newblock \href {https://aclanthology.org/C16-1291} {Neural machine translation
  with supervised attention}.
\newblock In \emph{Proceedings of {COLING} 2016, the 26th International
  Conference on Computational Linguistics: Technical Papers}, pages 3093--3102,
  Osaka, Japan. The COLING 2016 Organizing Committee.

\bibitem[{Liu et~al.(2020)Liu, Gu, Goyal, Li, Edunov, Ghazvininejad, Lewis, and
  Zettlemoyer}]{liu-etal-2020-multilingual-denoising}
Yinhan Liu, Jiatao Gu, Naman Goyal, Xian Li, Sergey Edunov, Marjan
  Ghazvininejad, Mike Lewis, and Luke Zettlemoyer. 2020.
\newblock \href {https://doi.org/10.1162/tacl_a_00343} {Multilingual denoising
  pre-training for neural machine translation}.
\newblock \emph{Transactions of the Association for Computational Linguistics},
  8:726--742.

\bibitem[{Lu et~al.(2018)Lu, Keung, Ladhak, Bhardwaj, Zhang, and
  Sun}]{lu-etal-2018-neural}
Yichao Lu, Phillip Keung, Faisal Ladhak, Vikas Bhardwaj, Shaonan Zhang, and
  Jason Sun. 2018.
\newblock \href {https://doi.org/10.18653/v1/W18-6309} {A neural interlingua
  for multilingual machine translation}.
\newblock In \emph{Proceedings of the Third Conference on Machine Translation:
  Research Papers}, pages 84--92, Brussels, Belgium. Association for
  Computational Linguistics.

\bibitem[{Luong et~al.(2015)Luong, Pham, and
  Manning}]{luong-etal-2015-effective}
Thang Luong, Hieu Pham, and Christopher~D. Manning. 2015.
\newblock \href {https://doi.org/10.18653/v1/D15-1166} {Effective approaches to
  attention-based neural machine translation}.
\newblock In \emph{Proceedings of the 2015 Conference on Empirical Methods in
  Natural Language Processing}, pages 1412--1421, Lisbon, Portugal. Association
  for Computational Linguistics.

\bibitem[{Ott et~al.(2019)Ott, Edunov, Baevski, Fan, Gross, Ng, Grangier, and
  Auli}]{ott-etal-2019-fairseq}
Myle Ott, Sergey Edunov, Alexei Baevski, Angela Fan, Sam Gross, Nathan Ng,
  David Grangier, and Michael Auli. 2019.
\newblock \href {https://doi.org/10.18653/v1/N19-4009} {fairseq: A fast,
  extensible toolkit for sequence modeling}.
\newblock In \emph{Proceedings of the 2019 Conference of the North {A}merican
  Chapter of the Association for Computational Linguistics (Demonstrations)},
  pages 48--53, Minneapolis, Minnesota. Association for Computational
  Linguistics.

\bibitem[{Ott et~al.(2018)Ott, Edunov, Grangier, and
  Auli}]{ott-etal-2018-scaling}
Myle Ott, Sergey Edunov, David Grangier, and Michael Auli. 2018.
\newblock \href {https://doi.org/10.18653/v1/W18-6301} {Scaling neural machine
  translation}.
\newblock In \emph{Proceedings of the Third Conference on Machine Translation:
  Research Papers}, pages 1--9, Brussels, Belgium. Association for
  Computational Linguistics.

\bibitem[{Pan et~al.(2021)Pan, Wang, Wu, and Li}]{pan-etal-2021-contrastive}
Xiao Pan, Mingxuan Wang, Liwei Wu, and Lei Li. 2021.
\newblock \href {https://doi.org/10.18653/v1/2021.acl-long.21} {Contrastive
  learning for many-to-many multilingual neural machine translation}.
\newblock In \emph{Proceedings of the 59th Annual Meeting of the Association
  for Computational Linguistics and the 11th International Joint Conference on
  Natural Language Processing (Volume 1: Long Papers)}, pages 244--258, Online.
  Association for Computational Linguistics.

\bibitem[{Papineni et~al.(2002)Papineni, Roukos, Ward, and
  Zhu}]{papineni-etal-2002-bleu}
Kishore Papineni, Salim Roukos, Todd Ward, and Wei-Jing Zhu. 2002.
\newblock \href {https://doi.org/10.3115/1073083.1073135} {{B}leu: a method for
  automatic evaluation of machine translation}.
\newblock In \emph{Proceedings of the 40th Annual Meeting of the Association
  for Computational Linguistics}, pages 311--318, Philadelphia, Pennsylvania,
  USA. Association for Computational Linguistics.

\bibitem[{Post(2018)}]{post-2018-call}
Matt Post. 2018.
\newblock \href {https://doi.org/10.18653/v1/W18-6319} {A call for clarity in
  reporting {BLEU} scores}.
\newblock In \emph{Proceedings of the Third Conference on Machine Translation:
  Research Papers}, pages 186--191, Brussels, Belgium. Association for
  Computational Linguistics.

\bibitem[{Shen et~al.(2018)Shen, Zhou, Long, Jiang, Pan, and
  Zhang}]{shen-etal-2018-disan}
Tao Shen, Tianyi Zhou, Guodong Long, Jing Jiang, Shirui Pan, and Chengqi Zhang.
  2018.
\newblock \href {https://doi.org/10.1609/aaai.v32i1.11941} {Disan: Directional
  self-attention network for rnn/cnn-free language understanding}.
\newblock \emph{Proceedings of the AAAI Conference on Artificial Intelligence},
  32(1).

\bibitem[{Siddhant et~al.(2020)Siddhant, Bapna, Cao, Firat, Chen, Kudugunta,
  Arivazhagan, and Wu}]{siddhant-etal-2020-leveraging}
Aditya Siddhant, Ankur Bapna, Yuan Cao, Orhan Firat, Mia Chen, Sneha Kudugunta,
  Naveen Arivazhagan, and Yonghui Wu. 2020.
\newblock \href {https://doi.org/10.18653/v1/2020.acl-main.252} {Leveraging
  monolingual data with self-supervision for multilingual neural machine
  translation}.
\newblock In \emph{Proceedings of the 58th Annual Meeting of the Association
  for Computational Linguistics}, pages 2827--2835, Online. Association for
  Computational Linguistics.

\bibitem[{Srivastava et~al.(2014)Srivastava, Hinton, Krizhevsky, Sutskever, and
  Salakhutdinov}]{JMLR:v15:srivastava14a}
Nitish Srivastava, Geoffrey Hinton, Alex Krizhevsky, Ilya Sutskever, and Ruslan
  Salakhutdinov. 2014.
\newblock Dropout: A simple way to prevent neural networks from overfitting.
\newblock \emph{JMLR}.

\bibitem[{Sutskever et~al.(2014)Sutskever, Vinyals, and
  Le}]{sutskever-etal-2014-sequence}
Ilya Sutskever, Oriol Vinyals, and Quoc~V. Le. 2014.
\newblock Sequence to sequence learning with neural networks.
\newblock In \emph{NeurIPS}.

\bibitem[{Szegedy et~al.(2016)Szegedy, Vanhoucke, Ioffe, Shlens, and
  Wojna}]{szegedy2016rethinking}
Christian Szegedy, Vincent Vanhoucke, Sergey Ioffe, Jon Shlens, and Zbigniew
  Wojna. 2016.
\newblock Rethinking the inception architecture for computer vision.
\newblock In \emph{CVPR}.

\bibitem[{Tan et~al.(2019)Tan, Ren, He, Qin, and
  Liu}]{tan-etal-2018-multilingual}
Xu~Tan, Yi~Ren, Di~He, Tao Qin, and Tie-Yan Liu. 2019.
\newblock \href {https://openreview.net/forum?id=S1gUsoR9YX} {Multilingual
  neural machine translation with knowledge distillation}.
\newblock In \emph{International Conference on Learning Representations}.

\bibitem[{Vaswani et~al.(2017)Vaswani, Shazeer, Parmar, Uszkoreit, Jones,
  Gomez, Kaiser, and Polosukhin}]{vaswani-etal-2017-attention}
Ashish Vaswani, Noam Shazeer, Niki Parmar, Jakob Uszkoreit, Llion Jones,
  Aidan~N. Gomez, Lukasz Kaiser, and Illia Polosukhin. 2017.
\newblock Attention is all you need.
\newblock In \emph{NeurIPS}.

\bibitem[{Wang et~al.(2020{\natexlab{a}})Wang, Tsvetkov, and
  Neubig}]{wang-etal-2020-balancing}
Xinyi Wang, Yulia Tsvetkov, and Graham Neubig. 2020{\natexlab{a}}.
\newblock \href {https://doi.org/10.18653/v1/2020.acl-main.754} {Balancing
  training for multilingual neural machine translation}.
\newblock In \emph{Proceedings of the 58th Annual Meeting of the Association
  for Computational Linguistics}, pages 8526--8537, Online. Association for
  Computational Linguistics.

\bibitem[{Wang et~al.(2020{\natexlab{b}})Wang, Zhai, and
  Hassan}]{wang-etal-2020-multi}
Yiren Wang, ChengXiang Zhai, and Hany Hassan. 2020{\natexlab{b}}.
\newblock \href {https://doi.org/10.18653/v1/2020.emnlp-main.75} {Multi-task
  learning for multilingual neural machine translation}.
\newblock In \emph{Proceedings of the 2020 Conference on Empirical Methods in
  Natural Language Processing (EMNLP)}, pages 1022--1034, Online. Association
  for Computational Linguistics.

\bibitem[{Wei et~al.(2019)Wei, Huang, Wang, Dai, and
  Chen}]{wei-etal-2019-online}
Hao-Ran Wei, Shujian Huang, Ran Wang, Xin-yu Dai, and Jiajun Chen. 2019.
\newblock \href {https://doi.org/10.18653/v1/N19-1192} {Online distilling from
  checkpoints for neural machine translation}.
\newblock In \emph{Proceedings of the 2019 Conference of the North {A}merican
  Chapter of the Association for Computational Linguistics: Human Language
  Technologies, Volume 1 (Long and Short Papers)}, pages 1932--1941,
  Minneapolis, Minnesota. Association for Computational Linguistics.

\bibitem[{Wu et~al.(2021)Wu, Li, Zhang, Li, Haffari, and
  Liu}]{wu-etal-2021-uncertainty}
Minghao Wu, Yitong Li, Meng Zhang, Liangyou Li, Gholamreza Haffari, and Qun
  Liu. 2021.
\newblock \href {https://doi.org/10.18653/v1/2021.emnlp-main.580}
  {Uncertainty-aware balancing for multilingual and multi-domain neural machine
  translation training}.
\newblock In \emph{Proceedings of the 2021 Conference on Empirical Methods in
  Natural Language Processing}, pages 7291--7305, Online and Punta Cana,
  Dominican Republic. Association for Computational Linguistics.

\bibitem[{Wu et~al.(2016)Wu, Schuster, Chen, Le, Norouzi, Macherey, Krikun,
  Cao, Gao, Macherey, Klingner, Shah, Johnson, Liu, Kaiser, Gouws, Kato, Kudo,
  Kazawa, Stevens, Kurian, Patil, Wang, Young, Smith, Riesa, Rudnick, Vinyals,
  Corrado, Hughes, and Dean}]{wu-etal-2016-googles}
Yonghui Wu, Mike Schuster, Zhifeng Chen, Quoc~V. Le, Mohammad Norouzi, Wolfgang
  Macherey, Maxim Krikun, Yuan Cao, Qin Gao, Klaus Macherey, Jeff Klingner,
  Apurva Shah, Melvin Johnson, Xiaobing Liu, Lukasz Kaiser, Stephan Gouws,
  Yoshikiyo Kato, Taku Kudo, Hideto Kazawa, Keith Stevens, George Kurian,
  Nishant Patil, Wei Wang, Cliff Young, Jason Smith, Jason Riesa, Alex Rudnick,
  Oriol Vinyals, Greg Corrado, Macduff Hughes, and Jeffrey Dean. 2016.
\newblock \href {http://arxiv.org/abs/1609.08144} {Google's neural machine
  translation system: Bridging the gap between human and machine translation}.
\newblock \emph{CoRR}, abs/1609.08144.

\bibitem[{Xia et~al.(2017)Xia, Tian, Wu, Lin, Qin, Yu, and
  Liu}]{xia-etal-2017-deliberation}
Yingce Xia, Fei Tian, Lijun Wu, Jianxin Lin, Tao Qin, Nenghai Yu, and Tie-Yan
  Liu. 2017.
\newblock \href
  {https://proceedings.neurips.cc/paper/2017/file/c6036a69be21cb660499b75718a3ef24-Paper.pdf}
  {Deliberation networks: Sequence generation beyond one-pass decoding}.
\newblock In \emph{Advances in Neural Information Processing Systems},
  volume~30. Curran Associates, Inc.

\bibitem[{Xie et~al.(2021)Xie, Feng, Gu, and Yu}]{xie-etal-2021-importance}
Wanying Xie, Yang Feng, Shuhao Gu, and Dong Yu. 2021.
\newblock \href {https://doi.org/10.18653/v1/2021.acl-long.445}
  {Importance-based neuron allocation for multilingual neural machine
  translation}.
\newblock In \emph{Proceedings of the 59th Annual Meeting of the Association
  for Computational Linguistics and the 11th International Joint Conference on
  Natural Language Processing (Volume 1: Long Papers)}, pages 5725--5737,
  Online. Association for Computational Linguistics.

\bibitem[{Yang et~al.(2019)Yang, Xie, Su, and Yuille}]{8953762}
C.~Yang, L.~Xie, C.~Su, and A.~L. Yuille. 2019.
\newblock Snapshot distillation: Teacher-student optimization in one
  generation.
\newblock In \emph{CVPR}.

\bibitem[{Zhang et~al.(2021{\natexlab{a}})Zhang, Bapna, Sennrich, and
  Firat}]{zhang-etal-2021-share}
Biao Zhang, Ankur Bapna, Rico Sennrich, and Orhan Firat. 2021{\natexlab{a}}.
\newblock \href {https://openreview.net/forum?id=Wj4ODo0uyCF} {Share or not?
  learning to schedule language-specific capacity for multilingual
  translation}.
\newblock In \emph{International Conference on Learning Representations}.

\bibitem[{Zhang et~al.(2020)Zhang, Williams, Titov, and
  Sennrich}]{zhang-etal-2020-improving}
Biao Zhang, Philip Williams, Ivan Titov, and Rico Sennrich. 2020.
\newblock \href {https://doi.org/10.18653/v1/2020.acl-main.148} {Improving
  massively multilingual neural machine translation and zero-shot translation}.
\newblock In \emph{Proceedings of the 58th Annual Meeting of the Association
  for Computational Linguistics}, pages 1628--1639, Online. Association for
  Computational Linguistics.

\bibitem[{Zhang et~al.(2019)Zhang, Song, Gao, Chen, Bao, and
  Ma}]{Zhang_2019_ICCV}
Linfeng Zhang, Jiebo Song, Anni Gao, Jingwei Chen, Chenglong Bao, and Kaisheng
  Ma. 2019.
\newblock Be your own teacher: Improve the performance of convolutional neural
  networks via self distillation.
\newblock In \emph{ICCV}.

\bibitem[{Zhang et~al.(2021{\natexlab{b}})Zhang, Meng, Tong, and
  Zhou}]{zhang-etal-2021-competence-based}
Mingliang Zhang, Fandong Meng, Yunhai Tong, and Jie Zhou. 2021{\natexlab{b}}.
\newblock \href {https://doi.org/10.18653/v1/2021.findings-emnlp.212}
  {Competence-based curriculum learning for multilingual machine translation}.
\newblock In \emph{Findings of the Association for Computational Linguistics:
  EMNLP 2021}, pages 2481--2493, Punta Cana, Dominican Republic. Association
  for Computational Linguistics.

\bibitem[{Zhou et~al.(2021)Zhou, Levy, Li, Ghazvininejad, and
  Neubig}]{zhou-etal-2021-distributionally}
Chunting Zhou, Daniel Levy, Xian Li, Marjan Ghazvininejad, and Graham Neubig.
  2021.
\newblock \href {https://doi.org/10.18653/v1/2021.emnlp-main.458}
  {Distributionally robust multilingual machine translation}.
\newblock In \emph{Proceedings of the 2021 Conference on Empirical Methods in
  Natural Language Processing}, pages 5664--5674, Online and Punta Cana,
  Dominican Republic. Association for Computational Linguistics.

\bibitem[{Zhou et~al.(2019)Zhou, Zhang, and Zong}]{zhou-etal-2019-synchronous}
Long Zhou, Jiajun Zhang, and Chengqing Zong. 2019.
\newblock \href {https://doi.org/10.1162/tacl_a_00256} {Synchronous
  bidirectional neural machine translation}.
\newblock \emph{Transactions of the Association for Computational Linguistics},
  7:91--105.

\bibitem[{Zhu et~al.(2020)Zhu, Yu, Cheng, and Luo}]{zhu-etal-2020-language}
Changfeng Zhu, Heng Yu, Shanbo Cheng, and Weihua Luo. 2020.
\newblock \href {https://doi.org/10.18653/v1/2020.acl-main.150} {Language-aware
  interlingua for multilingual neural machine translation}.
\newblock In \emph{Proceedings of the 58th Annual Meeting of the Association
  for Computational Linguistics}, pages 1650--1655, Online. Association for
  Computational Linguistics.

\bibitem[{Zhu et~al.(2021)Zhu, Feng, Zhao, Wang, and
  Li}]{zhu-etal-2021-counter-interference}
Yaoming Zhu, Jiangtao Feng, Chengqi Zhao, Mingxuan Wang, and Lei Li. 2021.
\newblock \href {https://doi.org/10.18653/v1/2021.findings-emnlp.240}
  {Counter-interference adapter for multilingual machine translation}.
\newblock In \emph{Findings of the Association for Computational Linguistics:
  EMNLP 2021}, pages 2812--2823, Punta Cana, Dominican Republic. Association
  for Computational Linguistics.

\end{thebibliography}
\bibliographystyle{acl_natbib}

\clearpage
\appendix

\appendix


\section{Data Statistics}
\label{section:data statistics}
Data statistics of the TED-8-Diverse and TED-8-Related are listed in Table~\ref{tab:data statistics.}.
Data statistics of the WMT dataset is listed in Table~\ref{tab:data statistics WMT.}.
\begin{table}[th]
\small
\renewcommand\tabcolsep{3.0pt}
\centering

\begin{tabular}{lc|lc}
\toprule
\multicolumn{2}{c|}{\textsc{Diverse}} & \multicolumn{2}{c}{\textsc{Related}}  \\
language & \#num & language & \#num \\
\midrule
\texttt{bos} (Bosnian) & 5,664 & 
\texttt{bel} (Belarusian) & 4,509 \\

\texttt{mar} (Marathi) & 9,840 & 
\texttt{aze} (Azerbaijani) & 5,946 \\

\texttt{hin} (Hindi) & 18,798 & 
\texttt{glg} (Glacian) & 10,017 \\

\texttt{mkd} (Macedonian) & 25,335 & 
\texttt{slk} (Slovak) & 61,470 \\

\texttt{ell} (Greek) & 134,327 & 
\texttt{cse} (Czech) & 103,093 \\

\texttt{bul} (Bulgarian) & 174,444 & 
\texttt{tur} (Turkish) & 182,470 \\

\texttt{fra} (French) & 192,304 & 
\texttt{por} (Portuguese) & 184,755 \\

\texttt{kor} (Korean) & 205,640 & 
\texttt{rus} (Russian) & 208,458 \\
\bottomrule
\end{tabular}
\caption{Data statistics for the TED-8-Diverse dataset and the TED-8-Related dataset. `\#num' refers to the number of sentence pairs in the training set.}
\label{tab:data statistics.}
\end{table}
\begin{table}[th]
\small
\renewcommand\tabcolsep{3.0pt}
\centering

\begin{tabular}{l|ccc}
\toprule
Language & Data Source & \#num \\
\midrule

\texttt{tr} (Turkish) & WMT17 & 5,000 \\

\texttt{ro} (Romanian) & WMT16 & 10,000 \\

\texttt{et} (Estonian) & WMT18 & 80,000 \\

\texttt{zh} (Chinese) & WMT17 & 400,000 \\

\texttt{de} (German) & WMT14 & 1,500,000 \\

\texttt{fr} (French) & WMT14 & 3,000,000 \\

\bottomrule
\end{tabular}
\caption{Data statistics for the WMT dataset. `\#num' refers to the number of sentence pairs in the training set.}
\label{tab:data statistics WMT.}
\end{table}

\section{\label{subsection:hyperparameters}Hyperparameters}

In this section, we list the details of hyperparameters we use for the experiments.
\begin{itemize}
    \item We adopt the architecture with 6 layers and 8 attention heads.
    \item The embedding dimension is 512 and the FFN has a dimension of 2048.
    \item We use Adam optimizer~\citep{DBLP:journals/corr/KingmaB14} with $\beta_1=0.9,\beta_2=0.98$, and the same learning rate schedule as \newcite{vaswani-etal-2017-attention}.
    \item Batch size is set to 64K and half-precision training is adopted~\cite{ott-etal-2018-scaling}. 
    \item For regularization, we use the dropout as 0.3~\cite{JMLR:v15:srivastava14a} and the label smoothing as 0.1~\cite{szegedy2016rethinking}.
    \item For sampling strategy, we use temperature-based sampling~\cite{arivazhagan-etal-2019-massively} and set $\tau=1$ on the TED-8-Diverse and TED-8-Related. And we set $\tau=5$ on the WMT dataset as the it is more imbalanced.
    \item For inference, we use beam search with beam size 5.
\end{itemize}



\section{Bilingual vs. Multilingual}
\label{subsection: bilingual performance}
\begin{table*}[t]
\small
    \centering
    \begin{tabular}{l|l | cccccccc |c}
    \toprule
        & \textbf{Method} & \textbf{bos} & \textbf{mar} & \textbf{hin} & \textbf{mkd} & \textbf{ell} & \textbf{bul} & \textbf{fra}   & \textbf{kor} & \textbf{Avg.} \\
    \midrule
        \multirow{2}{*}{M2O} 
        & \textsc{Bilingual} & 5.64 & 3.35 & 9.90 & 19.66 & 37.00 & 38.85 & \textbf{40.67} & \textbf{19.25} & 21.79 \\
        & \textsc{Multilingual} & \textbf{25.78} & \textbf{11.25} & \textbf{24.30} & \textbf{33.38} & \textbf{38.40} & \textbf{39.34} & 40.43 & 19.15 & \textbf{29.00} \\
        
        \midrule
        
        \multirow{2}{*}{O2M} 
        & \textsc{Bilingual} & 3.89 & 2.47 & 8.16 & 13.98 & 30.97 & 34.21 & 38.53 & 7.70 & 17.49 \\
        & \textsc{Multilingual} & \textbf{17.04} & \textbf{4.96} & \textbf{15.99} & \textbf{25.34} & \textbf{33.27} & \textbf{36.31} & \textbf{40.81} & \textbf{9.08} & \textbf{22.85} \\
    \bottomrule
    \end{tabular}
    \caption{BLEU scores of bilingual models and the multilingual model on the TED-8-Diverse dataset.
    Languages are ordered increasingly by data size from left to right. Bold indicates the higher BLEU score. We can find that the \textsc{Multilingual} model consistently outperforms the \textsc{Bilingual} model on each language pair except for \texttt{fra}$\rightarrow$\texttt{eng} and \texttt{kor}$\rightarrow$\texttt{eng}.}
    \label{tab:bilingual results diverse}
    \vspace{-3mm}
\end{table*}
\begin{table*}[t]
\small
    \centering
    \begin{tabular}{l|l | cccccccc | c}
    \toprule
        & \textbf{Method} & \textbf{aze} & \textbf{bel}	& \textbf{glg} & \textbf{slk} & \textbf{tur} & \textbf{rus} & \textbf{por} & \textbf{ces}	& \textbf{Avg.}\\
    \midrule
        \multirow{2}{*}{M2O} 
        & \textsc{Bilingual} & 2.16 & 1.8 & 10.1 & 23.62 & 26.51 & 24.97 & 44.53 & 25.77 & 19.93 \\
        & \textsc{Multilingual} & \textbf{13.11} & \textbf{19.57} & \textbf{30.14} & \textbf{32.36} & \textbf{27.07} & \textbf{25.85} & \textbf{44.89} & \textbf{29.66} & \textbf{27.83}\\
        \midrule

        \multirow{2}{*}{O2M} 
        & \textsc{Bilingual} & 1.59 & 1.78 & 9.25 & 17.62 & 14.51 & 19.59 & 39.15 & 17.97 & 15.18 \\
        & \textsc{Multilingual} &\textbf{7.40} & \textbf{13.09} & \textbf{25.71} & \textbf{25.66} & \textbf{16.93} & \textbf{20.94} & \textbf{41.68} & \textbf{23.35} & \textbf{21.85} \\
        
    \bottomrule
    \end{tabular}
    \caption{BLEU scores of bilingual models and the multilingual model on the TED-8-related dataset.
    Languages are ordered increasingly by data size from left to right.
    Bold indicates the higher BLEU score.
    We can find that the \textsc{Multilingual} model consistently outperforms the \textsc{Bilingual} model on all language pairs.}
    \label{tab:bilingual results related}
    \vspace{-3mm}
\end{table*}
\begin{table*}[t]
\small
    \centering
    \begin{tabular}{l|l | cccccc | c}
    \toprule
        & \textbf{Method} & \textbf{Tr} & \textbf{Ro} & \textbf{Et} & \textbf{Zh} & \textbf{De} & \textbf{Fr}	& \textbf{Avg.} \\
    \midrule
        \multirow{2}{*}{M2O} 
        & \textsc{Bilingual} & 0.78 & 5.00 & 5.99 & 11.60 & \textbf{27.94} & \textbf{33.57} & 14.15 \\
        & \textsc{Multilingual} & \textbf{9.89} & \textbf{23.12} & \textbf{16.92} & \textbf{12.72} & 26.68 & 31.57 & \textbf{20.15}\\
        \midrule

        \multirow{2}{*}{O2M} 
        & \textsc{Bilingual} & 0.34 & 3.34 & 4.76 & 19.38 & \textbf{24.11} & \textbf{36.07} & 14.67  \\
        & \textsc{Multilingual} & \textbf{8.83} & \textbf{18.42} &\textbf{ 13.34} & \textbf{20.78} & 21.11 & 31.93 & \textbf{19.07}  \\
        
    \bottomrule
    \end{tabular}
    \caption{BLEU scores of bilingual models and the multilingual model on the WMT dataset.
    Languages are ordered increasingly by data size from left to right.
    Bold indicates the better performance. We can see that the \textsc{Bilingual} model far outperforms the \textsc{Multilingual} on the high-source \texttt{fr}$\leftrightarrow$\texttt{en} and \texttt{de}$\leftrightarrow$\texttt{en}.
    However, when the training data is limited, the \textsc{Bilingual} model lags far behind the \textsc{Multilingual}.
    }
    \label{tab:bilingual results wmt}
    \vspace{-3mm}
\end{table*}
We list the results of bilingual models and the multilingual model in Table~\ref{tab:bilingual results diverse}, \ref{tab:bilingual results related}, \ref{tab:bilingual results wmt}.

\appendix
\end{document}